%% file: main.tex
\documentclass[runningheads]{llncs}

\usepackage[T1]{fontenc}
\usepackage{lmodern}

\usepackage{graphicx}
\graphicspath{{figures/}}
\usepackage{caption}
\usepackage{subcaption}

\usepackage[section]{placeins}

\usepackage{booktabs}
\usepackage{array}
\usepackage{tabularx}
\newcolumntype{L}{>{\raggedright\arraybackslash}X}
\usepackage{siunitx}
\usepackage{amsmath}
\usepackage{amssymb}
\usepackage{mathtools}

\usepackage{xcolor}
\usepackage[hidelinks]{hyperref}

\let\savedthebibliography\thebibliography
\renewcommand{\thebibliography}[1]{%
  \savedthebibliography{#1}%
  \scriptsize
  \setlength{\itemsep}{0pt plus 0.5pt}%
  \setlength{\parsep}{0pt}%
}
\begin{document}

\title{Calibrating Artificial Guilt: Neurally Grounded Reward Shaping for Prosocial Multi-Agent Reinforcement Learning}
\titlerunning{Calibrating Artificial Guilt for MARL}

\author{%
Aaditya Mehta\inst{1}\orcidID{0009-0000-5739-1426} \and
Arya Shah\inst{2}\orcidID{0000-0002-2649-1835}%
}
\authorrunning{A. Mehta et al.}

\institute{%
Mahatma Gandhi International School, Ahmedabad, India\\
\email{aadityamehta.work@gmail.com}
\and
Indian Institute of Technology, Gandhinagar, India\\
\email{arya.shah@iitgn.ac.in}%
}

\maketitle

\input{sections/00_abstract}
\input{sections/01_introduction}

\input{sections/02_related_work}

\input{sections/03_methodology}
\input{sections/05_experiments}
\input{sections/06_results}
\input{sections/07_discussion}
\input{sections/08_conclusion}
\input{sections/09_credits}

\bibliographystyle{splncs04}
\bibliography{references}

\end{document}

%% file: sections/00_abstract.tex
\begin{abstract}
Cooperative multi-agent reinforcement learning often adds social terms to
individual rewards, yet the scale of those terms is usually chosen by hand.
We ask whether a guilt signal can instead be calibrated from human neural and
behavioural data and transferred to artificial agents. Using the public SoDec
responsibility fMRI dataset (40 participants), we fit a subject-fixed-effects
regression of momentary-happiness changes on outcome-type counts and recover a
guilt weight as the Partner-negative minus Social-negative contrast
($\hat{w}=1.118$, Cohen's $d=0.214$). We embed this weight in a two-agent
Social Lottery environment and train independent Proximal Policy Optimization
actor-critics under four shaping regimes: neurally calibrated, uniform
constant, zero (selfish), and a unit-coefficient oracle. Across 1{,}000
evaluation episodes per condition, the calibrated agents track the human
Social safe-choice rate most closely ($0.459$ vs.\ human $0.484$;
$\mathrm{KL}=0.0012$), while the other three conditions deviate by one to
three orders of magnitude in KL. Human neurobehavioural priors can therefore
act as quantitative constraints on prosocial reward shaping.
\keywords{Multi-agent RL \and Reward shaping \and Guilt \and
Prosocial behaviour \and Value alignment}
\end{abstract}

%% file: sections/01_introduction.tex
\section{Introduction}
\label{sec:intro}

Mixed-motive multi-agent reinforcement learning (MARL) is a common setting for
studying cooperation among learning agents
\cite{leibo2017sequential,dafoe2020cooperative}. Independent self-interested
learners often defect in social dilemmas, so researchers augment individual
rewards with social signals such as inequity aversion
\cite{hughes2018inequity,fehr1999inequity}, social influence
\cite{jaques2019socialinfluence}, or diverse social preferences
\cite{mckee2020diversity}. Potential-based reward shaping
\cite{ng1999shaping} gives a formal route for such bias without changing the
optimal policy class under stated conditions.

Two design choices usually remain heuristic. The shaping magnitude is tuned by
hand or by sweep, and the cooperative reference is often a normative ideal
such as full equality rather than measured human choice. Agents can then satisfy
a designer criterion while missing the distribution of real social decisions.
Human social cognition offers a sharper alternative: momentary happiness in
interactive tasks is predictable from simple outcome equations
\cite{rutledge2014happiness}, and interpersonal guilt reliably engages the
anterior insula \cite{chang2011guilt,yu2014interpersonal}.

\begin{figure}[!t]
\centering
\includegraphics[width=0.92\linewidth]{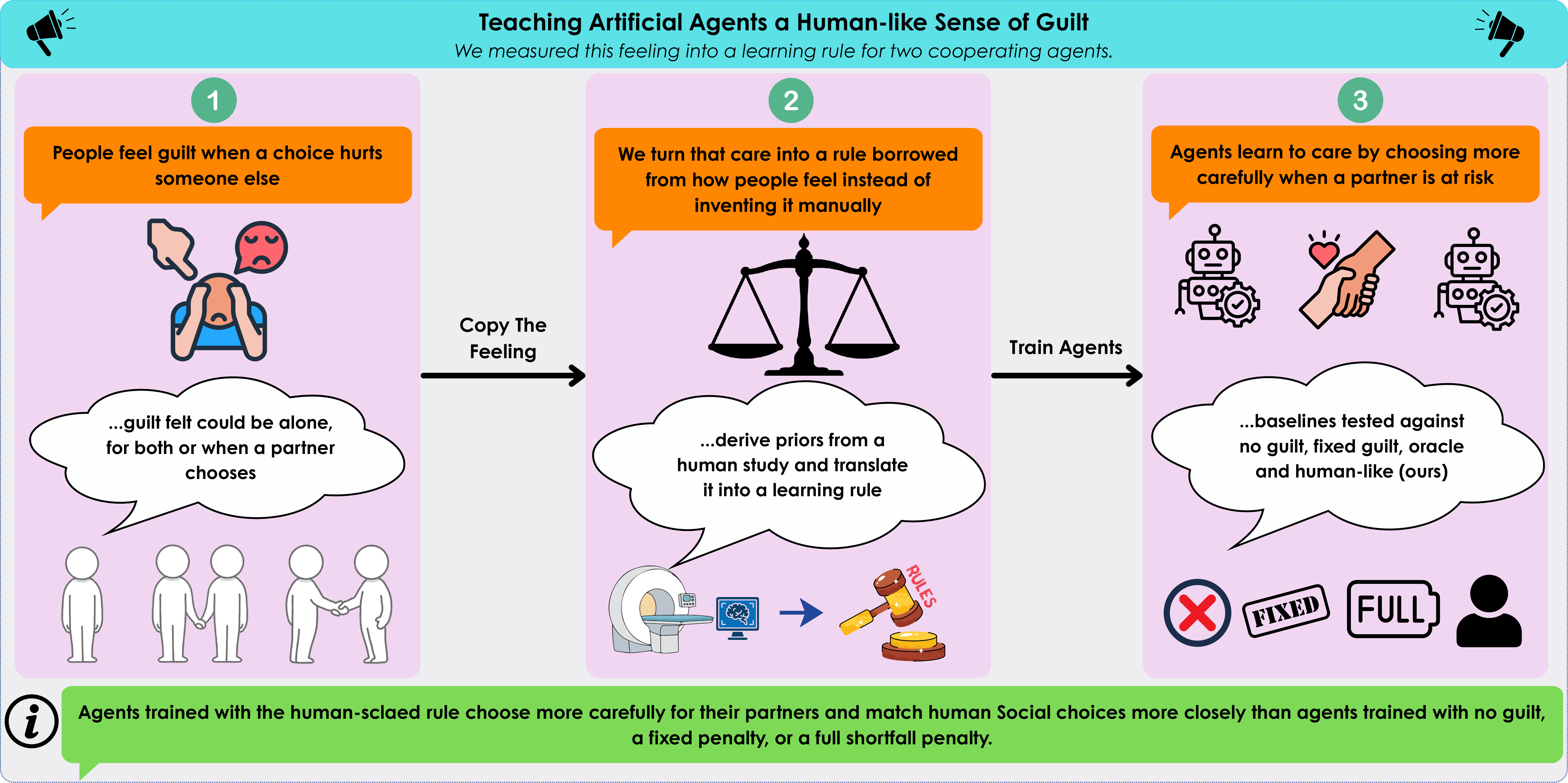}
\caption{Teaching artificial agents a human-like sense of guilt. People feel worse when their choice harms a partner; we turn that measured feeling into a learning rule for two cooperating agents. Agents trained with the human-scaled rule choose more carefully for their partners and match human Social choices more closely than agents trained with no guilt, a fixed penalty, or a full shortfall penalty.}
\label{fig:teaser}
\end{figure}

We calibrate an artificial guilt penalty from the SoDec responsibility dataset
(OpenNeuro ds005588) \cite{schultz2024responsibility,gadeke2025responsibility}.
Participants choose between a safe payoff and a risky lottery, either for
themselves alone (Solo) or for themselves and a partner (Social). A published
imaging analysis localises the interpersonal-guilt contrast to the left
anterior insula. Our pipeline turns the corresponding behavioural contrast into
a scalar weight $\hat{w}$ and injects it into independent Proximal Policy
Optimization (PPO) agents \cite{schulman2017ppo} in a matching Social Lottery
environment (Fig.~\ref{fig:teaser}).

The transfer needs no new pairwise preference labels and no inverse
reinforcement learning \cite{christiano2017preferences}. After training, the
neurally calibrated agents reach a safe rate of $0.459$ against a human Social
baseline of $0.484$, with
$\mathrm{KL}(P_{\mathrm{human}}\|P_{\mathrm{agent}})=0.0012$. That divergence
is an order of magnitude smaller than a unit-coefficient oracle penalty and
hundreds of times smaller than a selfish baseline. The calibrated coefficient
($\hat{w}=1.118$) is slightly larger than the oracle coefficient ($w=1$), yet
produces lower cumulative penalty mass because agents choose the safe option
more often. Magnitude, not only functional form, therefore carries alignment
information.

The main contributions of this work are: (1)~a calibration pipeline that turns
a published anterior-insula-indexed behavioural contrast into a single
transferable RL hyperparameter without extra human supervision;
(2)~a PettingZoo \cite{terry2021pettingzoo} Social Lottery environment that
reproduces the responsibility paradigm with lottery parameters sampled from
ds005588; and (3)~evidence that the scale of a human moral signal improves
behavioural alignment in MARL relative to designer-chosen constants.

%% file: sections/02_related_work.tex
\section{Related Work}
\label{sec:related}

We situate the work across five strands: cooperative MARL, reward shaping,
prosocial social signals, computational and neural accounts of guilt, and
alignment from human priors.

\paragraph{Cooperative and mixed-motive MARL.}
Sequential social dilemmas cast cooperation as a Markov game in which
independent learners often converge to defection
\cite{leibo2017sequential,perolat2017commonpool}, a failure mode that
cooperative AI agendas also emphasise \cite{dafoe2020cooperative}. Deep MARL
addresses coordination with centralised critics \cite{lowe2017maddpg},
opponent learning awareness \cite{foerster2018lola}, value decomposition
\cite{sunehag2017vdn,rashid2018qmix}, population game-theoretic solvers
\cite{lanctot2017psro}, and strong independent PPO baselines
\cite{yu2022mappo}. Large-scale multi-agent systems and surveys further map
these ideas under complex dynamics
\cite{vinyals2019starcraft,hernandez2019survey}. Across this literature,
success is usually defined by team return or equilibrium quality, not by match
to a measured human choice distribution.

\paragraph{Reward shaping and intrinsic motivation.}
Potential-based reward shaping biases learning without changing the optimal
policy class under stated conditions \cite{ng1999shaping,wiewiora2003potential}.
Intrinsic motivation supplies related additive signals for exploration and
skill discovery, including prediction-error curiosity
\cite{pathak2017curiosity}, random network distillation \cite{burda2019rnd},
and diversity-driven objectives \cite{eysenbach2019diayn}. We adopt the
additive shaping template, but treat the coefficient as an empirically
estimated social prior rather than an exploration bonus or hand-chosen
constant.

\paragraph{Social and prosocial reward signals.}
A parallel line injects explicitly social terms into MARL rewards: inequity
aversion \cite{hughes2018inequity} grounded in Fehr and Schmidt preferences
\cite{fehr1999inequity}, counterfactual social influence
\cite{jaques2019socialinfluence}, heterogeneous social preferences
\cite{mckee2020diversity}, reciprocity \cite{eccles2019reciprocity},
prosocial objectives \cite{peysakhovich2018prosocial,lerer2017maintaining},
evolved intrinsic rewards for altruism \cite{wang2019evolving}, and
human-centred mechanism design with deep RL \cite{koster2022democratic}.
Related behavioural models outside MARL include Rabin's fairness equilibrium
\cite{rabin1993fairness}, Bolton and Ockenfels ERC \cite{bolton2000erc}, and
Charness and Rabin social preferences \cite{charness2002social}. In these systems
the functional form of the social term is carefully motivated; the magnitude
is still typically swept. Our setting is orthogonal: fix a guilt-like form and
import its scale from human data.

\paragraph{Guilt, happiness, and the anterior insula.}
Behavioural economics treats guilt as an explicit utility component in
psychological games \cite{battigalli2007guilt,charness2006promises}. Momentary
happiness in interactive tasks is well approximated by linear functions of own
and partner outcomes \cite{rutledge2014happiness}. Neuroimaging localises
interpersonal guilt and related social-affective monitoring to the anterior
insula and connected prefrontal and temporal circuits
\cite{chang2011guilt,yu2014interpersonal,wagner2011guiltspecific,koban2013integration,zahn2009neural,moll2006charitable}.
The SoDec responsibility dataset and its associated analysis
\cite{schultz2024responsibility,gadeke2025responsibility}, shared via OpenNeuro
under BIDS with fMRIPrep derivatives
\cite{markiewicz2021openneuro,gorgolewski2016bids,esteban2019fmriprep},
provide a public behavioural and neural handle on that contrast. We use the
behavioural events to estimate a transferable scalar and do not re-analyse
BOLD maps here.

\paragraph{Alignment from human priors.}
Value alignment typically relies on inverse RL
\cite{ng2000inverse,abbeel2004apprenticeship}, cooperative IRL
\cite{hadfieldmenell2016cirl}, preference comparisons
\cite{christiano2017preferences}, scalable reward modelling
\cite{leike2018scalable}, RLHF \cite{ouyang2022instructgpt}, or
constitutional AI feedback \cite{bai2022constitutional}. A complementary
programme argues for structural priors drawn from neuroscience
\cite{hassabis2017neuroscience,botvinick2020deeprl,marblestone2016integration,wang2018prefrontal}.
Our differentiation is narrow: we calibrate one social shaping coefficient
from a published neurobehavioural contrast, instantiate it in a PettingZoo
Social Lottery environment \cite{terry2021pettingzoo} with independent PPO
learners \cite{schulman2017ppo}, and score agents against the same human task
distribution without collecting new preference labels.

%% file: sections/03_methodology.tex
\section{Methodology}
\label{sec:methodology}

The pipeline has three stages: extract a guilt weight from the SoDec
responsibility dataset, embed that weight in a two-agent Social Lottery
environment, and train independent PPO actor-critics under four shaping
regimes.

\subsection{Dataset}
\label{sec:methodology:dataset}

We use the SoDec responsibility dataset (OpenNeuro ds005588)
\cite{schultz2024responsibility,gadeke2025responsibility}, which contains
forty participants (ages 22 to 50) scanned across two runs of a responsibility
task ($\mathrm{TR}=2.5$\,s). On each trial, participants chose between a
deterministic safe payoff and a 50/50 risky lottery under Solo (self only),
Social (self and partner), or Partner (partner chooses for both) conditions.
Risky outcomes were drawn independently for the two players, and momentary
happiness was rated on a 0 to 100 scale every two to four trials. The original
imaging analysis localised the Social-versus-Partner happiness-decrement
contrast to the left anterior insula
\cite{gadeke2025responsibility}. Our pipeline uses only the behavioural event
records.

\subsection{Extracting the Guilt Weight}
\label{sec:methodology:extract}

For each participant and run we group events into \emph{happiness epochs}: the
sequence of outcomes between consecutive happiness ratings. Within each epoch
we count nine outcome types defined by the
$\{\text{self},\text{other}\} \times \{\text{pos},\text{neg}\}$ valence cross
the $\{\text{Solo},\text{Social},\text{Partner}\}$ condition factorial, plus
three deterministic safe-outcome types. Following Rutledge et al.\
\cite{rutledge2014happiness}, we regress the happiness change
$\Delta h_{i,k}$ for participant $i$ at epoch $k$ on these counts with subject
fixed effects:
\begin{equation}
\Delta h_{i,k}
  = \alpha + \sum_{j=1}^{9} \beta_{j}\,n_{j,k} + \gamma_{i} + \varepsilon_{i,k},
\label{eq:happiness-regression}
\end{equation}
where $n_{j,k}$ is the count of outcome type $j$ and $\gamma_{i}$ is a
participant intercept (drop-one dummy coding). We estimate
Eq.~\eqref{eq:happiness-regression} by ordinary least squares via the
pseudo-inverse, which remains stable under sparse outcome counts.

\begin{figure}[!t]
\centering
\begin{subfigure}[t]{0.48\linewidth}
\centering
\includegraphics[width=\linewidth]{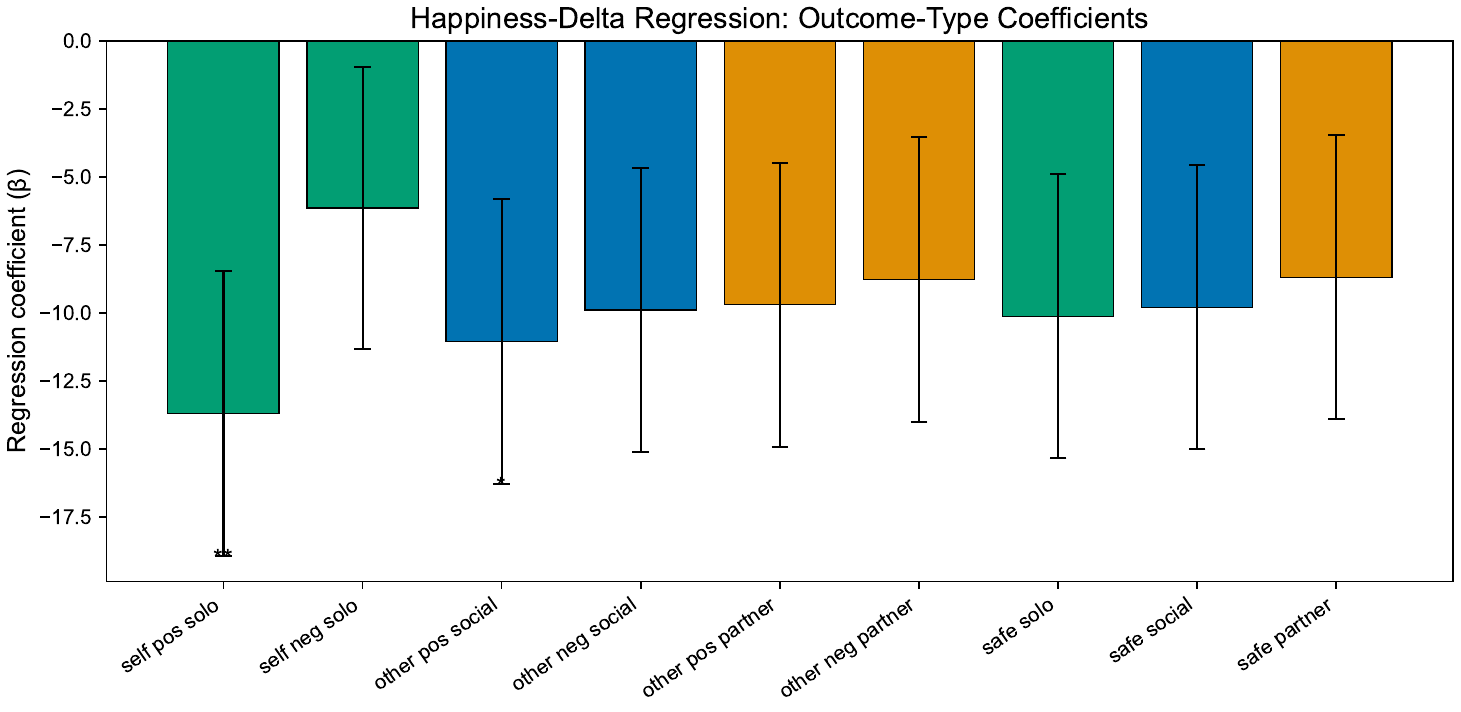}
\caption{Coefficients $\hat{\beta}_{j}$ ($\pm 1$ SE).}
\label{fig:guilt-regression-coefficients}
\end{subfigure}
\hfill
\begin{subfigure}[t]{0.48\linewidth}
\centering
\includegraphics[width=\linewidth]{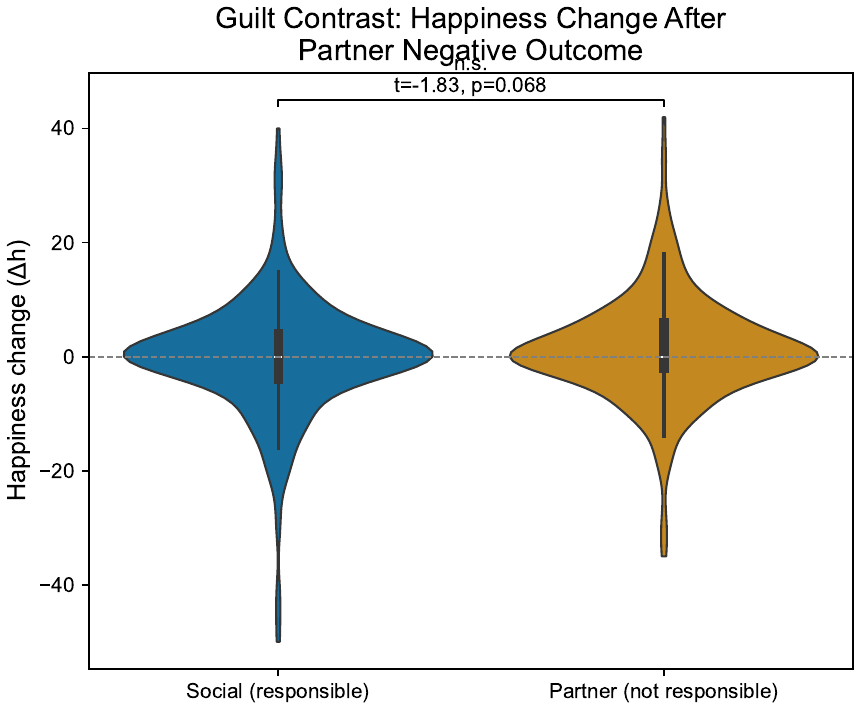}
\caption{$\Delta h$ after one partner-negative outcome.}
\label{fig:guilt-contrast-violin}
\end{subfigure}
\caption{Extracting the neural guilt weight from ds005588.}
\label{fig:guilt-extraction}
\end{figure}

Interpersonal guilt in this paradigm is the extra happiness decrement after a
partner's negative outcome when the participant was responsible (Social),
relative to when the partner was responsible (Partner):
\begin{equation}
w \;\coloneqq\;
\beta_{\mathrm{other\_neg\_partner}}
- \beta_{\mathrm{other\_neg\_social}}.
\label{eq:guilt-weight}
\end{equation}
On ds005588 we obtain $\hat{w}=1.118$ with pooled standard error
$\sigma_{\hat{w}}=7.39$, Cohen's $d=0.214$, and one-sided $p=0.44$. The point
estimate has the predicted sign with a small-to-moderate standardised effect,
but the behavioural sample alone does not reach classical significance. We
therefore treat $\hat{w}$ as a prior on penalty magnitude, not as a
stand-alone hypothesis test. Figure~\ref{fig:guilt-extraction} shows the fitted
coefficients and the contrast distribution.

\subsection{Social Lottery Environment}
\label{sec:methodology:env}

\begin{figure}[!t]
\centering
\includegraphics[width=0.72\linewidth]{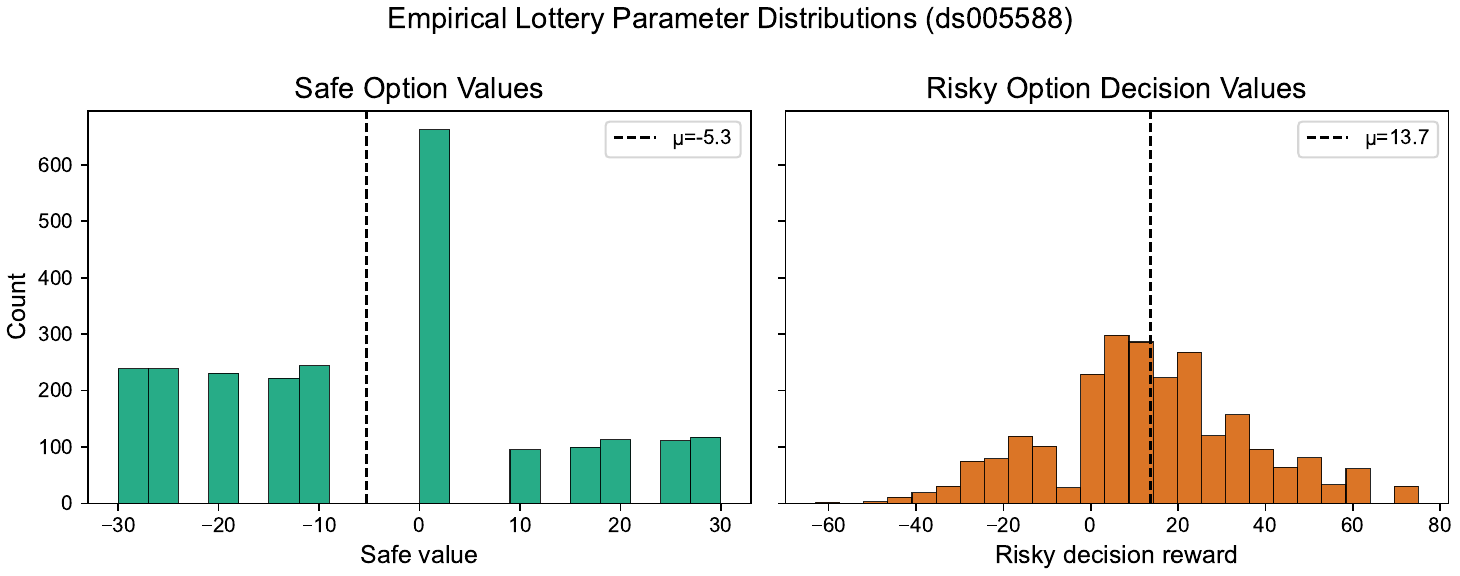}
\caption{Empirical safe-option and risky expected-value distributions in
ds005588; environment lotteries are sampled from these.}
\label{fig:lottery-distributions}
\end{figure}

We cast the paradigm as a two-agent Markov game
$\mathcal{M}=\langle \mathcal{N},\mathcal{S},\{\mathcal{A}_{i}\},P,\{R_{i}\},T\rangle$
implemented as a PettingZoo ParallelEnv \cite{terry2021pettingzoo} on the
Gymnasium API \cite{towers2024gymnasium}. Agents
$\mathcal{N}=\{0,1\}$ alternate as \emph{decider} across $T=20$ rounds per
episode. Both observe the same four-dimensional vector
\begin{equation}
o_{t} =
\bigl(v_{\mathrm{safe}}/c,\;
      v_{\mathrm{high}}/c,\;
      v_{\mathrm{low}}/c,\;
      \mathbb{1}\{\mathrm{decider}\}\bigr)^{\!\top},
\label{eq:observation}
\end{equation}
with normaliser $c=30$. Actions are binary,
$\mathcal{A}_{i}=\{\textsc{Safe},\textsc{Risky}\}$; only the decider's action
is consequential. We sample $v_{\mathrm{safe}}$ and the risky expected value
from the empirical marginals in ds005588
(Fig.~\ref{fig:lottery-distributions}) and draw the risky half-spread from
$|\mathcal{N}(15,5)|$.

Let $r_{d},r_{p}$ be round payoffs to decider and partner, and let
$\mathrm{EV}=(v_{\mathrm{high}}+v_{\mathrm{low}})/2$. The partner shortfall
after a risky choice is $\Delta_{p}=\max(0,\mathrm{EV}-r_{p})$. Shaped rewards
are
\begin{align}
R_{d}(s,a)
&= \tfrac{1}{c}\bigl(r_{d}-w\,\Delta_{p}\,
   \mathbb{1}\{a=\textsc{Risky}\}\bigr),
\label{eq:reward-decider}\\
R_{p}(s,a)
&= \tfrac{1}{c}\,r_{p}.
\label{eq:reward-partner}
\end{align}
The penalty applies only to the decider, and only when a risky action yields a
low partner outcome. Table~\ref{tab:conditions} lists the four values of $w$.

\begin{table}[!t]
\caption{Reward-shaping conditions. NeuroGuilt uses the fitted contrast; Oracle
applies a unit coefficient on the same shortfall.}
\label{tab:conditions}
\centering
\scriptsize
\setlength{\tabcolsep}{5pt}
\begin{tabular}{@{}lll@{}}
\toprule
\textbf{Condition} & \boldmath$w$ & \textbf{Interpretation} \\
\midrule
\textsc{NeuroGuilt} & $1.118$ & Eq.~\eqref{eq:guilt-weight} \\
\textsc{Uniform}    & $0.5$   & Data-agnostic constant \\
\textsc{Zero}       & $0$     & Selfish baseline \\
\textsc{Oracle}     & $1$     & Unit-coefficient shortfall penalty \\
\bottomrule
\end{tabular}
\end{table}

\subsection{Multi-Agent PPO Training}
\label{sec:methodology:ppo}

We train two independent PPO actor-critics \cite{schulman2017ppo}, one per
agent. Each network is a two-layer MLP trunk of $64$ $\tanh$ units with linear
policy and value heads. With GAE advantages $\hat{A}_{t}$ and returns
$\hat{R}_{t}$, the clipped objective is
\begin{equation}
\begin{aligned}
\mathcal{L}^{\mathrm{PPO}}(\theta)
&= \mathbb{E}_{t}\Bigl[
     \min\bigl(\rho_{t}\hat{A}_{t},\,
               \mathrm{clip}(\rho_{t},1-\epsilon,1+\epsilon)\hat{A}_{t}\bigr)
   \Bigr] \\
&\quad - c_{v}\,\mathbb{E}_{t}\bigl[(V_{\theta}(s_{t})-\hat{R}_{t})^{2}\bigr]
  + c_{e}\,\mathbb{E}_{t}\bigl[\mathcal{H}(\pi_{\theta}(\cdot|s_{t}))\bigr],
\end{aligned}
\label{eq:ppo}
\end{equation}
where $\rho_{t}=\pi_{\theta}(a_{t}|s_{t})/\pi_{\theta_{\mathrm{old}}}(a_{t}|s_{t})$.
Hyperparameters (Table~\ref{tab:hyperparams}) were chosen once on
\textsc{Zero} and held fixed across conditions so that differences isolate $w$.
Each condition trains for $20{,}000$ episodes with seed $42$ and is evaluated
greedily on $1{,}000$ episodes with seed $42{+}999$, so training and evaluation
draw on disjoint stochastic streams. End-to-end training takes on the order of
one hour per condition on a single CPU. We intentionally avoid condition-specific
learning-rate or entropy retuning: any change in safe rate or KL is then
attributable to $w$ rather than to optimiser asymmetry.

\begin{table}[!t]
\caption{Shared PPO hyperparameters for all four conditions.}
\label{tab:hyperparams}
\centering
\scriptsize
\setlength{\tabcolsep}{4pt}
\begin{tabular}{@{}llll@{}}
\toprule
\textbf{Param} & \textbf{Value} & \textbf{Param} & \textbf{Value} \\
\midrule
Episodes & $20{,}000$ & PPO epochs & $4$ \\
Rounds $T$ & $20$ & Mini-batch & $64$ \\
Hidden units & $64$ & Clip $\epsilon$ & $0.2$ \\
Optimiser & Adam & Entropy $c_{e}$ & $0.01$ \\
Learning rate & $3\times 10^{-4}$ & Value $c_{v}$ & $0.5$ \\
$\gamma$ & $0.99$ & Grad clip & $0.5$ \\
$\lambda$ (GAE) & $0.95$ & Eval episodes & $1{,}000$ \\
\bottomrule
\end{tabular}
\end{table}


%% file: sections/05_experiments.tex
\section{Experimental Design}
\label{sec:experiments}

After training, each condition is evaluated for $1{,}000$ greedy episodes under
identical environment dynamics and the shared hyperparameters of
Table~\ref{tab:hyperparams}. The design targets three questions:
(RQ1)~Does a neurally calibrated guilt penalty change learned policies relative
to a selfish baseline?
(RQ2)~How does $\hat{w}=1.118$ compare with designer coefficients
$w\in\{0.5,1\}$?
(RQ3)~Does the resulting choice distribution align more closely with human
Social behaviour than the baselines?

We report social welfare (SW; episode-summed return), safe rate (SR; fraction
of decider rounds choosing safe), reward inequality (IE; absolute difference in
agent returns), and guilt-penalty mass (GP). For RQ3 we compute KL divergence
on the binary safe/risky distribution against the human Social baseline
$P_{H,\mathrm{Social}}=0.484$ from ds005588. Fixed seeds keep the comparison
focused on the shaped reward in Eq.~\eqref{eq:reward-decider}.

%% file: sections/06_results.tex
\section{Results}
\label{sec:results}

We organise findings around the three research questions in
Sec.~\ref{sec:experiments}.

\begin{figure}[!t]
\centering
\begin{subfigure}[t]{0.48\linewidth}
\centering
\includegraphics[width=\linewidth]{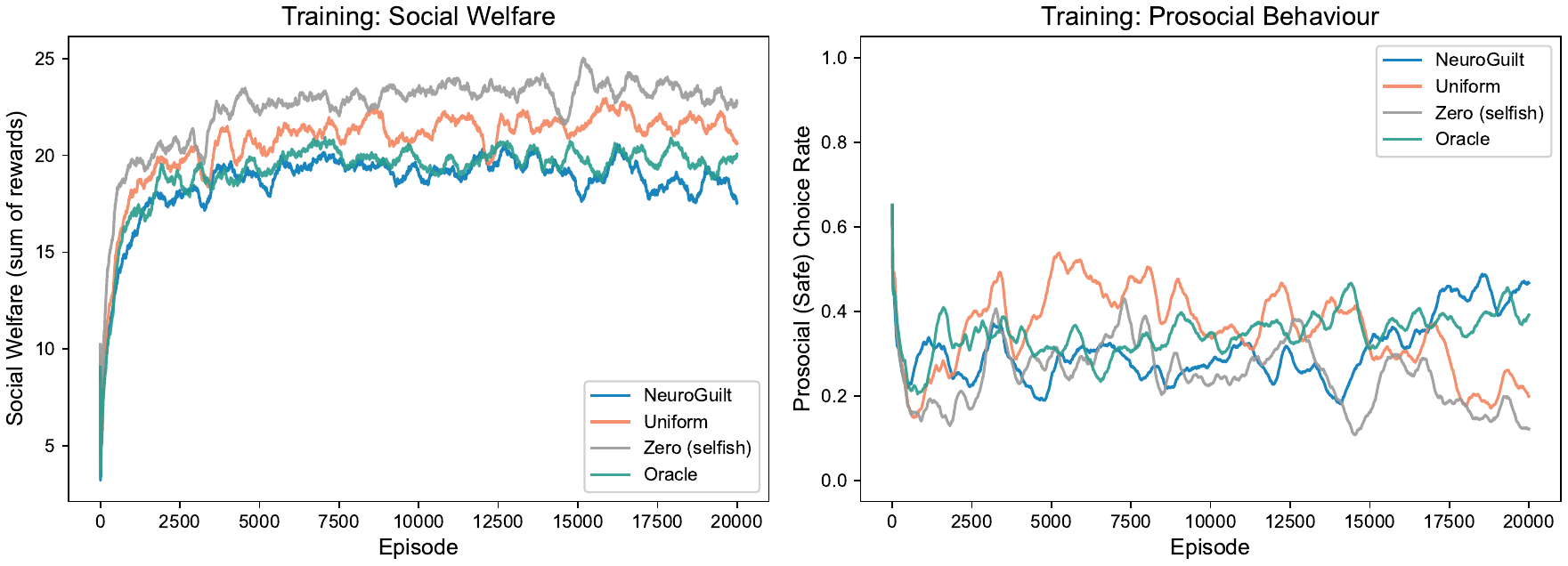}
\caption{Training curves.}
\label{fig:training-curves}
\end{subfigure}
\hfill
\begin{subfigure}[t]{0.48\linewidth}
\centering
\includegraphics[width=\linewidth]{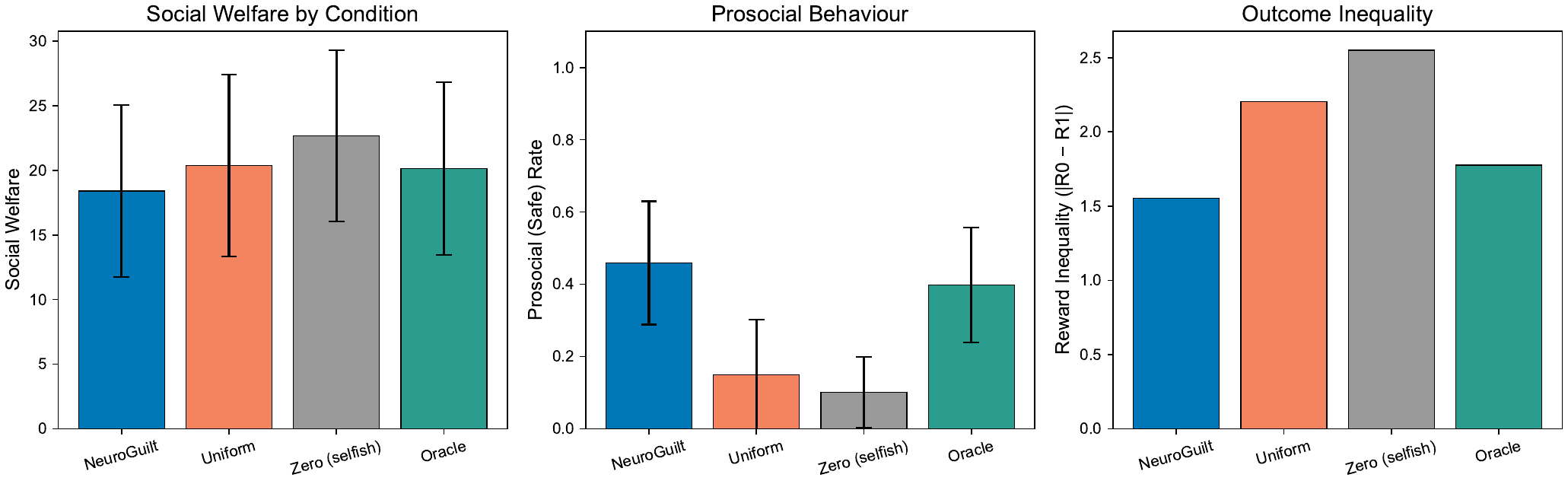}
\caption{Steady-state outcomes.}
\label{fig:condition-comparison}
\end{subfigure}
\caption{Training dynamics and group-level trade-offs across conditions.}
\label{fig:results-panel}
\end{figure}

\paragraph{Training dynamics and trade-offs (RQ1, RQ2).}
Figure~\ref{fig:results-panel}a shows social welfare and prosocial rate over
$20{,}000$ training episodes. Welfare curves separate early:
\textsc{Zero} climbs highest because it never penalises risky play. Safe-rate
trajectories separate more slowly. \textsc{Zero} and \textsc{Uniform} settle at
low safe rates, while \textsc{Oracle} and \textsc{NeuroGuilt} plateau near
$0.40$ and $0.46$. Table~\ref{tab:main-results} and
Fig.~\ref{fig:results-panel}b report steady-state metrics over $1{,}000$
evaluation episodes. Most metrics move with effective guilt exposure, with one
notable pattern: \textsc{NeuroGuilt} records lower reward inequality ($1.554$)
than \textsc{Oracle} ($1.776$) and lower cumulative guilt mass ($2.863$ vs.\
$3.105$), even though its coefficient is larger ($1.118$ vs.\ $1$). Safer
policies trigger the penalty less often, so cumulative mass is not a monotone
readout of $w$.

\begin{table}[!t]
\caption{Steady-state metrics over $1{,}000$ greedy episodes (mean $\pm$ SD
where applicable). SW: social welfare; SR: safe rate; IE: inequality; GP:
guilt-penalty mass.}
\label{tab:main-results}
\centering
\scriptsize
\setlength{\tabcolsep}{3.5pt}
\begin{tabular}{@{}lccccc@{}}
\toprule
\textbf{Condition}
& \textbf{SW}
& \textbf{SR}
& \textbf{IE}
& \textbf{GP}
& \textbf{KL($P_{H}\|P_{A}$)} \\
\midrule
\textsc{Zero}
& $22.669\pm 6.621$ & $0.101\pm 0.098$ & $2.549$ & $0.000$ & $0.4717$ \\
\textsc{Uniform}
& $20.378\pm 7.042$ & $0.150\pm 0.152$ & $2.203$ & $2.146$ & $0.3097$ \\
\textsc{Oracle}
& $20.135\pm 6.702$ & $0.398\pm 0.159$ & $1.776$ & $3.105$ & $0.0153$ \\
\textbf{\textsc{NeuroGuilt}}
& $\mathbf{18.400\pm 6.652}$
& $\mathbf{0.459\pm 0.171}$
& $\mathbf{1.554}$
& $\mathbf{2.863}$
& $\mathbf{0.0012}$ \\
\midrule
Human (Social) & n/a & $0.484$ & n/a & n/a & n/a \\
\bottomrule
\end{tabular}
\end{table}

\begin{figure}[!t]
    \centering
    \begin{minipage}{0.48\textwidth}
        \centering
        \includegraphics[width=\linewidth]{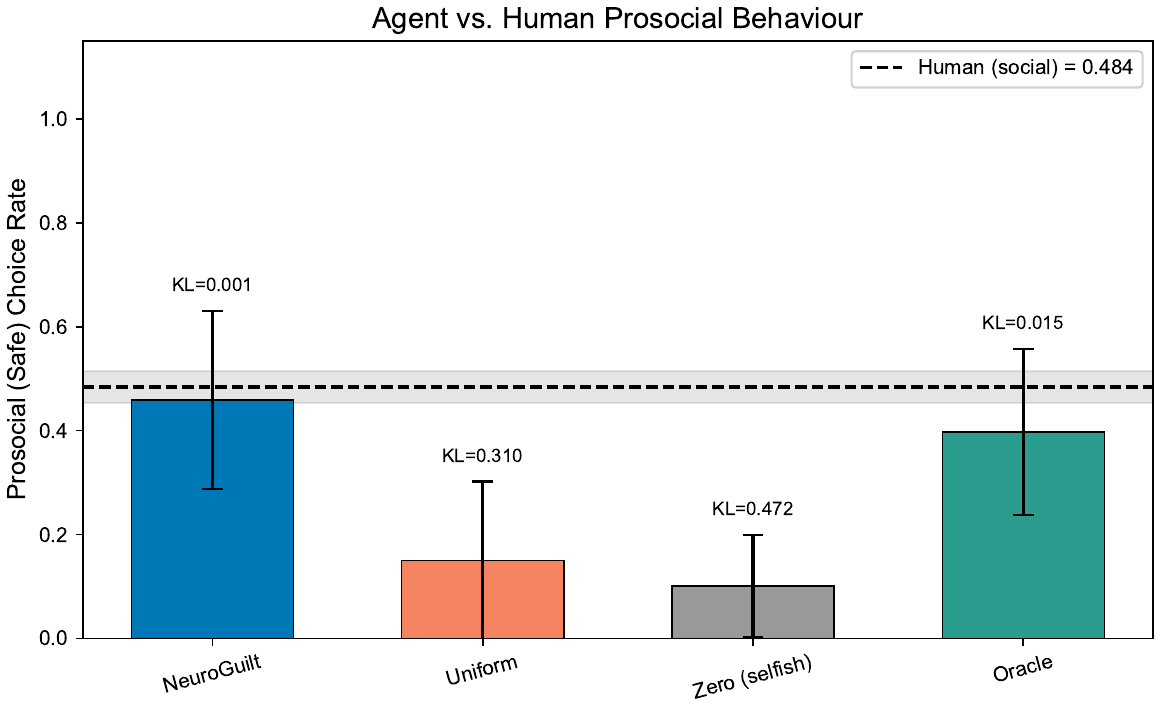}
        \caption{Human Social safe rate (dashed) versus agent safe rates under the four shaping conditions. Annotated values are $\mathrm{KL}(P_{\mathrm{human}}\|P_{\mathrm{agent}})$.}
        \label{fig:human-vs-agent}
    \end{minipage}\hfill
    \begin{minipage}{0.48\textwidth}
        \centering
        \includegraphics[width=\linewidth]{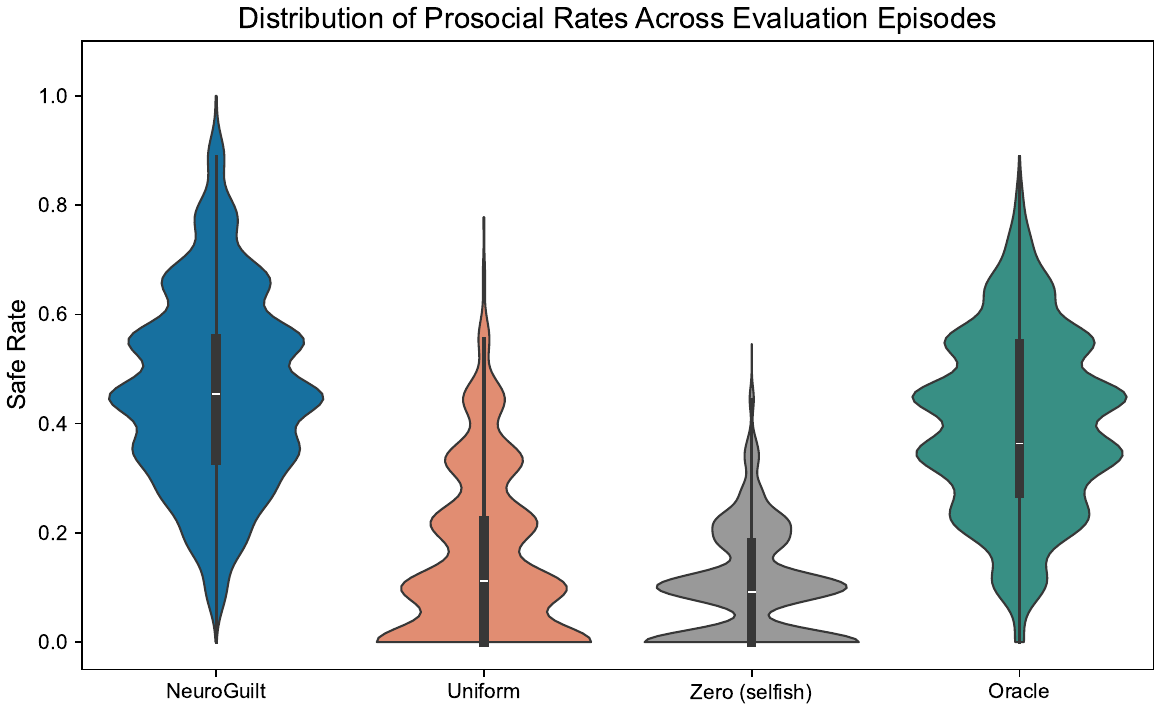}
        \caption{Per-episode prosocial (safe) choice-rate distributions.}
        \label{fig:safe-rate-distributions}
    \end{minipage}
\end{figure}

\paragraph{Alignment with the human distribution (RQ3).}
The KL column of Table~\ref{tab:main-results} is the headline comparison.
\textsc{NeuroGuilt}'s safe rate ($0.459$) lies within $0.025$ of the human
Social rate ($0.484$). In KL terms, its divergence is $12.8\times$ smaller than
\textsc{Oracle}'s and $393\times$ smaller than \textsc{Zero}'s
(Fig.~\ref{fig:human-vs-agent}). The calibrated magnitude is therefore closer
to human choice than either a weaker constant or a unit-coefficient shortfall
penalty. Per-episode safe-rate distributions
(Fig.~\ref{fig:safe-rate-distributions}) show the same pattern:
\textsc{Oracle} and \textsc{NeuroGuilt} retain wider support that resembles
human Social variability, whereas the low-guilt baselines collapse toward
near-deterministic risky play.

%% file: sections/07_discussion.tex
\section{Discussion}
\label{sec:discussion}

The central empirical claim is that a single scalar taken from a public
neurobehavioural contrast can constrain prosocial MARL more tightly than
common designer defaults. Relative to the promises in Sec.~\ref{sec:intro}, the
results deliver on behavioural alignment: \textsc{NeuroGuilt} recovers a human-like
safe rate and a KL divergence in the $10^{-3}$ regime without new preference
labels. They also qualify a simple ``stronger penalty is better'' intuition.
\textsc{Oracle} uses $w=1$ on the same partner shortfall; \textsc{NeuroGuilt}
uses $\hat{w}=1.118$. The calibrated coefficient is not smaller, yet it yields
higher safe rate, lower inequality, lower cumulative penalty mass, and much
lower KL to humans. Alignment here tracks the \emph{data-derived scale}, not a
monotone ranking of coefficient size.

Two limitations matter for interpretation. First, $\hat{w}$ is estimated from
behavioural happiness regressions with a non-significant classical test
($p=0.44$). We treat it as a prior transferred into RL, not as a confirmed
neural effect size; subject-level anterior-insula betas from the accompanying
imaging derivatives remain unused. Second, the Social Lottery is a short-horizon
abstraction of the scanner task. It preserves responsibility structure and
empirical lottery values, but not delayed credit assignment, communication, or
richer mixed-motive grids used elsewhere in MARL. Claims should stay inside
that scope: magnitude calibration helps when the environment matches the human
paradigm closely enough for the safe/risky distribution to be a meaningful
target.

The broader implication is modest but concrete. Preference-based alignment and
neurobehavioural calibration need not compete. The former collects new human
judgments; the latter reuses measured social priors already published with open
datasets. For moral signals with an established neural and behavioural
signature, that reuse can replace at least one layer of hand tuning.

A practical reading for MARL design follows. When a social penalty has a clear
human operationalisation, sweep less and measure more: fit the scale on the
same paradigm you evaluate against, hold the optimiser fixed, and report
divergence to the human distribution rather than only welfare or cooperation
rate. Welfare alone would have preferred \textsc{Zero} in our tables; the
alignment metrics reverse that ranking. That separation is the reason to keep
human choice as an explicit target.

%% file: sections/08_conclusion.tex
\section{Conclusion}
\label{sec:conclusion}

This work contributed a calibration path from a published interpersonal-guilt
contrast to a transferable reward weight for multi-agent reinforcement learning,
a Social Lottery environment matched to the SoDec responsibility paradigm, and
evidence that the scale of that weight improves alignment with human Social
choices. With $\hat{w}=1.118$, independent PPO agents reached a safe rate of
$0.459$ against a human baseline of $0.484$ and a KL divergence of $0.0012$,
outperforming selfish, constant, and unit-coefficient baselines on the same
task. Replacing the behavioural proxy with subject-specific neural betas, and
testing the same magnitude-transfer idea in longer-horizon dilemmas, are the
natural next steps enabled by the existing dataset.

%% file: sections/09_credits.tex
\begin{credits}
\subsubsection{\discintname}
The authors have no competing interests to declare that are relevant to the
content of this article.
\end{credits}